\documentclass[sigconf,authorversion, screen,nonacm]{acmart}

\author{Liyuan Zhang}
\affiliation{%
\institution{Chongqing University}
\city{Chongqing}
\country{China}
}
\email{lyzhang@cqu.edu.cn}
\author{Yuhang Zhou}
\affiliation{%
\institution{Chongqing University}
\city{Chongqing}
\country{China}
}
\email{yuhangzhou@cqu.edu.cn}
\author{Lei Zhang$^{(}$\textsuperscript{\Letter}$^)$}
\affiliation{%
\institution{Chongqing University}
\city{Chongqing}
\country{China}
}
\email{leizhang@cqu.edu.cn}
\AtBeginDocument{%
  \providecommand\BibTeX{{%
    \normalfont B\kern-0.5em{\scshape i\kern-0.25em b}\kern-0.8em\TeX}}}

\usepackage[misc,geometry]{ifsym}
\usepackage{threeparttable}
\usepackage{graphicx}
\usepackage{amsfonts}
\usepackage{amsthm,amsmath}
\usepackage{mathrsfs}
\usepackage{multirow}
\usepackage{pdfpages}
\begin{document}

%%
%% The "title" command has an optional parameter,
%% allowing the author to define a "short title" to be used in page headers.
\title{On the Robustness of Domain Adaption to Adversarial Attacks}

\begin{abstract}

State-of-the-art deep neural networks (DNNs) have been proved to have excellent performance on unsupervised domain adaption (UDA). However, recent work shows that DNNs perform poorly when being attacked by adversarial samples, where these attacks are implemented by simply adding small disturbances to the original images. Although plenty of work has focused on this, as far as we know, there is no systematic research on the robustness of unsupervised domain adaption model.
 Hence, we discuss the robustness of unsupervised domain adaption against adversarial attacking for the first time. We benchmark various settings of adversarial attack and defense in domain adaption, and propose a cross domain attack method based on pseudo label. Most importantly, we analyze the impact of different datasets, models, attack methods and defense methods. Directly, our work proves the limited robustness of unsupervised domain adaptation model, and we hope our work may facilitate the community to pay more attention to improve the robustness of the model against attacking.
\end{abstract}

%%
%% The code below is generated by the tool at http://dl.acm.org/ccs.cfm.
%% Please copy and paste the code instead of the example below.
%%

\begin{comment}
\begin{CCSXML}
<ccs2012>
 <concept>
  <concept_id>10010520.10010553.10010562</concept_id>
  <concept_desc>Computer systems organization~Embedded systems</concept_desc>
  <concept_significance>500</concept_significance>
 </concept>
 <concept>
  <concept_id>10010520.10010575.10010755</concept_id>
  <concept_desc>Computer systems organization~Redundancy</concept_desc>
  <concept_significance>300</concept_significance>
 </concept>
 <concept>
  <concept_id>10010520.10010553.10010554</concept_id>
  <concept_desc>Computer systems organization~Robotics</concept_desc>
  <concept_significance>100</concept_significance>
 </concept>
 <concept>
  <concept_id>10003033.10003083.10003095</concept_id>
  <concept_desc>Networks~Network reliability</concept_desc>
  <concept_significance>100</concept_significance>
 </concept>
</ccs2012>
\end{CCSXML}

\ccsdesc[500]{Computer systems organization~Embedded systems}
\ccsdesc[300]{Computer systems organization~Redundancy}
\ccsdesc{Computer systems organization~Robotics}
\ccsdesc[100]{Networks~Network reliability}

\end{comment}

%%
%% This command processes the author and affiliation and title
%% information and builds the first part of the formatted document.
\maketitle

\section{Introduction}
The powerful representation ability makes DNNs perform better and better in domain adaption. Great effort has been devoted to developing robust domain adaption models \cite{ganin2015unsupervised,2015Domain,2018Transferable,2017Simultaneous,long2017deep,2018Uns}, in order to overcome the huge shift among domains. For example, the accuracy of the state-of-the-art result of domain adaption  model on the Office-31 datasets \cite{2010Adapting} is 86.6\% \cite{long2017conditional}, increasing rapidly from 34\% when the dataset was first released in 2010.

\begin{figure}[h]

  \centering
  \includegraphics[width=\linewidth]{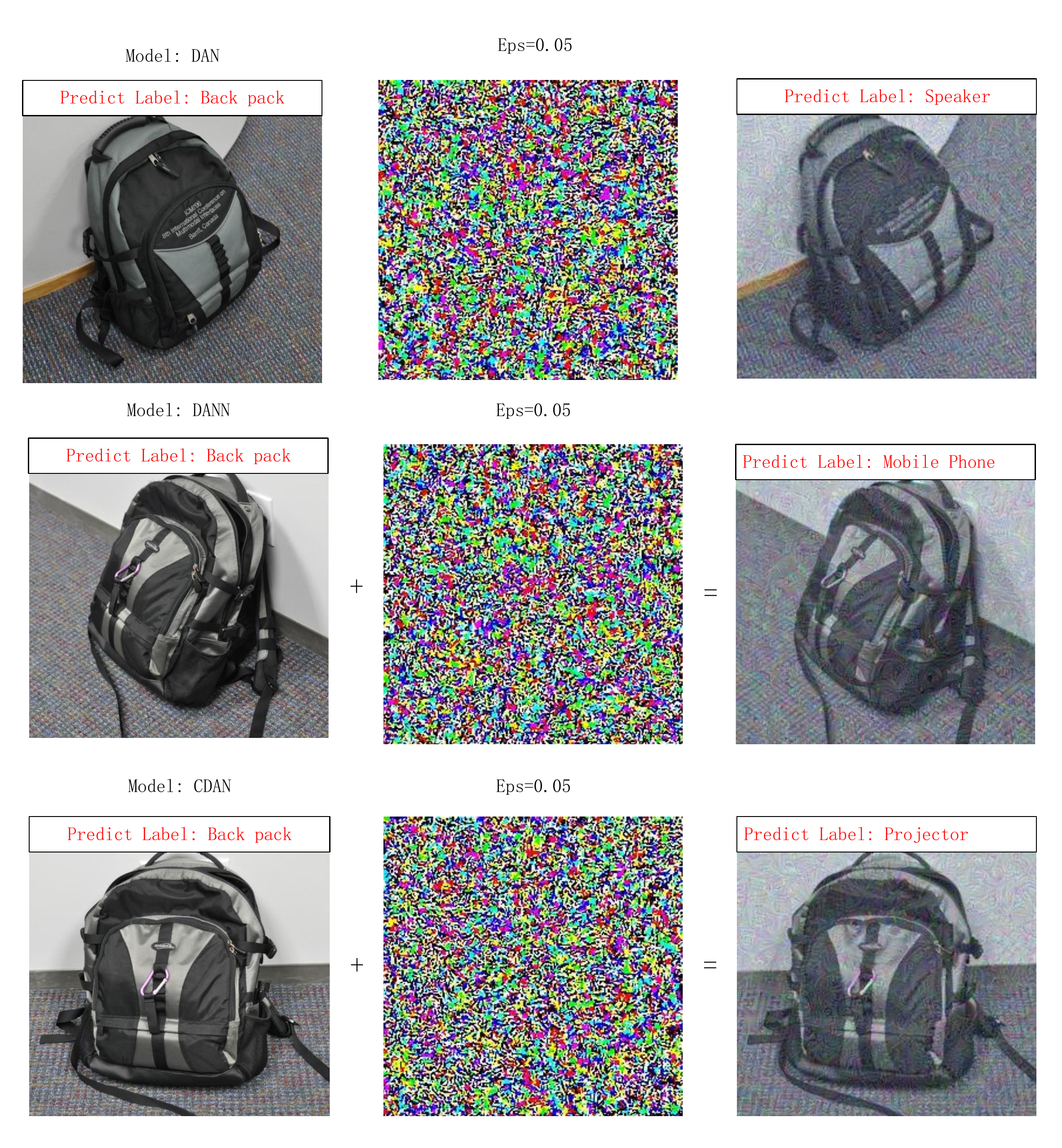}
  \caption{Adversarial attack examples. After being added the noise, which does not affect human's discrimination, the backpack is misclassified by model as speaker, mobil phone and projector.}
  \label{fig:1}
\end{figure}

Yet, \cite{2013Intriguing} proposed that DNN can be very vulnerable to adversarial attacks \cite{2013Evasion,2004Adversarial1,2013Intriguing}. Fig.\ref{fig:1} shows an adversarial case, where a backpack image from Office-31 is presented. By adding noise to the samples, the network will mistakenly think that the backpack is a speaker, a mobile phone or a projector.
This phenomenon has attracted a lot of attention. Numerous strategies have been proposed to get more robust models to adversarial examples \cite{2014Explaining,2016Adversarial,2017Towards1}. However, current works almost focus on the standard image classification models (Inception in \cite{2016Adversarial}, and GoogleNet, VGG and ResNet in \cite{2016Delving}) using small datasets like mnist and cifar-10, $etc$. Hence, these works can hardly play guiding role to more concrete and complex tasks,  and the vulnerability of modern DNNs to adversarial attacks on more complex systems remains unclear, such as unsupervised domain adaption covering different domains. Besides, existing attack methods usually require the label information to generate the adversarial samples \cite{2014Explaining,2017Towards1,2017One}, which is unrealistic for domain adaption {since the labels of the target domain is not available}.

Therefore, it is meaningful to explore the robustness of the model against adversarial attacks in more complex domain adaption systems, and a new method to attack domain adaption model is required. In this paper, we present, what to our knowledge is, the first systematic research on unsupervised domain adaption model's robustness to adversarial attacks. Our main contributions can be summarized as follows:

(1) We first define and benchmark various experimental settings
for {Adversarial Domain Attack and Defense (ADAD)}, including white-box \cite{2014Explaining,2017Towards1,2017Adversarial} and black-box \cite{2020DaST, 2020AdvFlow} attack, non-targeted and targeted attack, white-box and black-box defense, $etc$. And we first choose three different methods to explore their robustness on three datasets of different scales including small, medium and large;

(2) We first propose a Fast Gradient Sign Method based on Pseudo Label (PL-FGSM), which can be used as a basic attack method for domain adaptation. {This method can attack the target domain samples by assigning pseudo labels to them.} Extensive experiments show that this method is effective for domain adaptation model;

(3) We first extensively study the robustness of different existing domain adaption models to adversarial samples, and compare different attack settings against the same model. Besides, the defense capability of different defense paradigms is also discussed. {Finally, we will discuss some interesting phenomena that may exist only in domain adaption.}

Some relevant works to ours are \cite{0On1} about  semantic segmentation, \cite{2019Metric} about person re-identification, and \cite{2017Towards1} about face detection. We hope that our work can facilitate the development of robust feature learning, and the research of adversarial attack and defense of domain adaption.

\section{RELATED WORK}
\subsection{Domain Adaption}
Transferring knowledge from a source domain with sufficient supervision to an unlabeled target domain is an important yet challenging problem. Unsupervised domain adaptation (UDA) addresses the challenge by learning a model which can use in different domains of different distributions.
In shallow architectures, domain adaptation is mainly achieved by matching the marginal distributions \cite{sugiyama2007direct, pan2010domain, gong2013connecting} or the conditional distributions \cite{courty2017joint, zhang2013domain}.
In recent years, unsupervised domain adaptation has made remarkable advances in deep learning architectures and the strategies can be divided into two main categories.

The first category uses statistical discrepancy metrics to measure the distance across domains, and reduces domain shift by constraining statistics \cite{long2015learning,long2017deep,li2020deep,zhang2019bridging,pan2019transferrable,cui2020towards,li2020enhanced}. For example, DDC \cite{tzeng2014deep} uses linear-kernel maximum mean discrepancies (MMD) \cite{sejdinovic2013equivalence} to adapt a single layer in order to maximize domain invariance. The later work of DAN \cite{long2015learning} addressed these problems by using multi-kernel MMD \cite{gretton2012kernel, gretton2012optimal} in multiple task-specific layers. Further, JAN \cite{long2017deep} explores joint MMD to enforce joint distribution alignment between domains. Based on the optimal transport (OT) distance \cite{zhang2019optimal, li2020deep} learn the optimal transport plan by enhanced transport distance (ETD).

The second category learns domain-invariant features in the confrontation between domain classifier and feature extractor \cite{saito2018maximum, wang2019transferable, ganin2015unsupervised, long2017conditional, li2019joint}. For example, DANN \cite{ganin2015unsupervised} introduces domain discriminator to domain adaptation. Further, CDAN \cite{long2017conditional} proposes a conditional domain discriminator. Such discriminator conditions the class information with feature representation, which can match the discrepancy of different distributions better. Besides, MCD \cite{saito2018maximum} employs a new adversarial strategy where the adversarial manner occurs between the feature extractor and classifiers rather than the feature extractor and the domain discriminator.

Although these models have achieved better and better performance, their robustness to adversarial samples has not been fully explored.

\subsection{Adversarial Attack and Defense}
Since the discovery of adversarial examples for DNNs \cite{2013Intriguing}, plenty of attacking methods have been proposed in CV community. Adversarial samples aim at fooling the network while making the noise imperceptible to human beings. \cite{2014Explaining} proposes to generate adversarial examples by using a single step based on the sign of the gradient for each pixel; \cite{2016Adversarial} implements the above method in an iterative way; \cite{moosavi2016deepfool} attacks the deep classifier by finding the nearest classification boundary.  \cite{papernot2016limitations} utilize the Jacobian matrix to implicitly conduct a fixed length of noise through the direction of each axis. \cite{2017One} proposes to modify the single-pixel adversarial attack. In addition, some methods for visualizing noise are also proposed. \cite{2017Adversarial} shows a way to generate special glasses, which can be used to successfully achieve directional or non directional attacks.  \cite{2017Adversarial} proposes to overlay a visible adversarial patch on the image and successfully fool the deep network. Adversarial attack is also discussed in some more complex tasks such as person re-identification \cite{2020Transferable, 2020Adversarial}, and semantic segmentation \cite{0On1} and so on. However, as discussed Section 1, the above methods all require the label information to generate the adversarial samples which is unrealistic for domain adaptation.

 For defense, \cite{2014Explaining} proposes to add confrontation samples in the defense process, which means that the generated adversarial samples are added to the training set as new training samples, \cite{papernot2016distillation} mentions that in the distillation network, the gradient of student network is smoother. This feature can be used to resist the confrontation samples generated based on gradient. \cite{liao2018defense} proposes a denoiser network; \cite{2017Miti} discuss that the random change of pixel level may lead to the failure of these specific disturbance points in the adversarial samples. \cite{zheng2016improving} proposes to force the network to learn similar feature representation from the original sample and the adversarial sample of the same image.

However, the above defense methods only discuss the defense performance of standard classification tasks. Most experiments was only carried out on the small MNIST dataset, and some defense was not effective on CIFAR-10 \cite{0On1}, underlining the importance of testing on multiple datasets. \cite{2017Ensemble} also found that adversarially trained models are still susceptible to black-box attacks generated from other networks. Besides, as far as we know, there is no defense method that can effectively resist all attacks. Hence, these defense methods can't provide a exact reference for the the state-of-the-art networks we consider in this work.

\section{METHODOLOGY}
Our methodology consists of three components. In 3.1 we formulate the generalized domain adaptation model. And we propose the loss with pseudo labels in 3.2. In 3.3 and 3.4, we first benchmark various settings of domain adversarial attack and defense, and then propose our methods.

\subsection{Generalized Domain Adaptation}
In unsupervised domain adaptation (UDA) \cite{pan2016survey}, we are provided with a source domain
$
\mathcal{D}_{s}=\left\{\left(\mathbf{x}_{i}^{s}, y_{i}^{s}\right)\right\}_{i=1}^{n_{s}}
$
 with $n_s$ labeled examples, and a target domain $\mathcal{D}_{t}=\left\{\mathbf{x}_{j}^{t}\right\}_{j=1}^{n_{t}}$
with $n_t$ unlabeled examples. The source and target domains are sampled from distributions $p$ and $q$ respectively, and $p \neq q$.
The challenge of UDA is to train a classifier model $y=M(x)$ with a low target risk $R_{\mathcal{D}_{t}}(M)=\operatorname{Pr}_{(\mathbf{x}, \mathbf{y}) \sim \mathcal{D}_{t}}[M(\mathbf{x}) \neq \mathbf{y}]$. The model $M(\theta,w,b)$ consists of two components $F(\theta)$ and $G(w,b)$, where $F(\theta)$ is the feature extractor and $G(w,b)$ is the class predictor. The target risk can be bounded by the source risk $R_{\mathcal{D}_{s}}(M)=\operatorname{Pr}_{(\mathbf{x}, \mathbf{y}) \sim \mathcal{D}_{s}}[M(\mathbf{x}) \neq \mathbf{y}]$ plus the distribution discrepancy $\operatorname{disc}(\mathrm{p}, \mathrm{q})$.

To constraint lower risk of source domain, UDA method calculate the loss of the classifier on the source domain as:
\begin{equation}
\label{2.1.1}\mathcal{L}_{\text {classifer}} = \frac{1}{n_{s}} \sum_{\mathbf{x}_{i}^{s} \in \mathcal{D}_{s}} \mathcal{L}_{y}\left({G}\left(\mathbf{x}_{i}^{s}\right), \mathbf{y}_{i}^{s}\right)
\end{equation}
where $L_{y}$ is classifier loss.
To constraint the distribution discrepancy, the method based on information statistics metrics (DDC \cite{tzeng2014deep}, DAN \cite{long2015learning}, JAN \cite{long2017deep}, etc.) calculates the distance between source and target distributions:
\begin{equation}
\label{2.1.2}
\mathcal{L}_{\text {transfer }} = \ d \left(\mathcal{D}_{s},\mathcal{D}_{t}\right)
\end{equation}
And the method based on adversarial (DANN \cite{ganin2015unsupervised},CDAN \cite{long2017conditional},MCD \cite{saito2018maximum}, etc.) training uses domain discriminator loss as transfer loss:
\begin{equation}
\label{2.1.3}
\mathcal{L}_{\text {transfer}}=\sum_{                                                                    {x}_{i} \in \mathcal{D}_{s} \cup \mathcal{D}_{t}}\mathcal{L}_{\mathrm{d}}\left(\mathcal{D}\left(\mathrm{F}\left(\mathrm{x}_{\mathrm{i}}\right)\right), \mathrm{d}_{\mathrm{i}}\right)
\end{equation}
where $d_{i}$ is the domain label of $x_{i}$.

The general UDA loss can be  formulated as:
\begin{equation}
\label{2.1.4}
\mathcal{L}_{DA}=\mathcal{L}_{\text {classifer}}+\mathcal{L}_{\text {transfer }}
\end{equation}

We choose the classic DAN \cite{long2015learning}, DANN \cite{ganin2015unsupervised} and CDAN \cite{long2017conditional} models as the attacked models in UDA.

\subsection{Loss with Pseudo Labels}

Existing adversarial attacks methods, such as FGSM \cite{2014Explaining}, FGSM ll \cite{2016Adversarial}, I-FGSM and I-FGSM ll \cite{2016Adversarial,2017Towards1} algorithms are implemented under the standard classification task, where we need the ground truths of the samples to calculate the loss function, such as cross entropy loss. However, in domain adaptation, we are unable to get the label information of the target domain. So we propose the loss with pseudo labels (PL), i.e., $\mathcal{L}_{PL}$ for domain adversarial attack.

The cross-entropy loss with target pseudo labels $\hat{y}_{i}^{t}$ is:
\begin{equation}
\label{3.1.1}l_{cross\_entropy}^{t} = \frac{1}{n_{t}} \sum_{\mathbf{x}_{i}^{t} \in \mathcal{D}_{t}} \mathcal{L}_{y}\left(\mathrm{G}\left(\mathbf{x}_{i}^{t}\right), \hat{y}_{i}^{t}\right)
\end{equation}
So the final objective $\mathcal{L}_{PL}$ can be defined as:
\begin{equation}
\label{3.1.2}
\mathcal{L}_{PL}=\mathcal{L}_{\text{DA}}+l_{cross\_entropy}^{t}
\end{equation}
where $\mathcal{L}_{\text {DA}}$ is the domain adaptation loss in Eq.(\ref{2.1.4}).

%\subsection{Adversarial Domain Attack}
\subsection{Adversarial Domain Attack}

For a given classifier $G(w,b)$, the adversarial sample $x_{adv}$ is obtained by adding some disturbance $r$ to the original sample $x$. Such kind of disturbance $r$ may be imperceptible to human eyes. \cite{moosavi2016deepfool} formulates the adversarial attack task as a conditional optimization model:
\begin{equation}
\label{3.2.0}
\Delta(\boldsymbol{x} ; G):=\min _{\boldsymbol{r}}\|\boldsymbol{r}\|_{2}, \text { subject to } G(\boldsymbol{x}+\boldsymbol{r}) \neq G(\boldsymbol{x})
\end{equation}
where $x$ is an image and $G(x)$ is the predictor. We call $\Delta(\boldsymbol{x} ; G)$ the robustness of $G(w,b)$ at point $x$. Greater $\Delta(\boldsymbol{x} ; G)$  means that greater disturbance is required to destroy the model.

{We benchmark several different attack settings for domain adaption, including \textit{non-targeted attack}, \textit{targeted attack}, \textit{white-box attack} and \textit{black-box attack} in following part.}
\subsubsection{ Non-targeted Attack}
It tends to destroy the similarity between images of the same label, or push a sample away from the classification decision boundary until it becomes another class, i.e., the model is fooled to make wrong predictions:
\begin{equation}\label{3.2.1}G(x+r) \neq G(x)\end{equation}
The attacker does not care about which class the adversarial samples are assigned, as long as they are not the true class.

In the untargeted attack, to acquire a misclassified adversarial image, the optimization goal of attack is to maximize loss $\mathcal{L}_{PL}$, which can be defined as:
\begin{equation}\label{3.2.2}
\underset{\mathrm{x}_{\text {adv }}}{\max } \ \mathcal{L}_{\mathrm{PL}}\left(\mathrm{M}\left(\mathrm{x}_{\text {adv }} ; \theta, \mathrm{w}, \mathrm{b}\right), \mathrm{y}\right)
\end{equation}

\begin{figure*}[t]
  \centering
  %width=0.9\linewidth,height=0.37\linewidth
  \includegraphics[width=0.85\linewidth,height=0.37\linewidth]{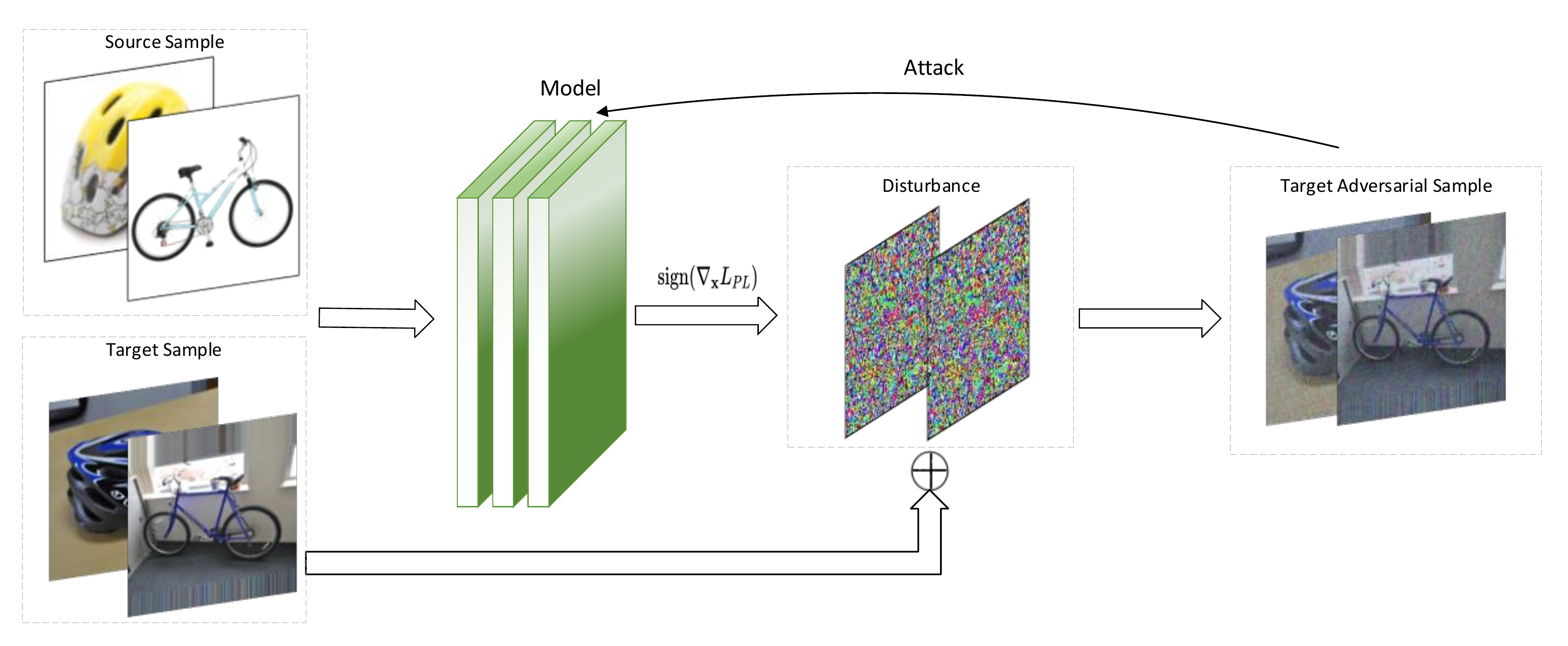}
    \caption{Process of PL-FGSM. The inputs to the model are samples of the source and target domain. $\nabla_{\mathbf{x}} \mathcal{L}_{P L}$is the gradient of $\mathcal{L}_{P L}$ on the sample x. The model is domain adaptation model which is trained by clean samples. We use source and target samples to calculate the loss $\mathcal{L}_{P L}$ and obtain disturbance $r$ by $\operatorname{sign}\left(\nabla_{\mathbf{x}} \mathcal{L}_{PL}\right)$. Then we add the disturbance to the clean images to get the adversarial samples and use adversarial samples to the attack the model.
}
  \label{fig:10}
\end{figure*}

We achieve this optimization objective by making iterative steps in the negative direction of gradient descent direction \(\operatorname{sign}\left(\nabla_{\mathbf{x}} \mathcal{L}_{P L}\left(M\left(\mathbf{X}_{n}^{\mathrm{adv}}\right), y\right)\right) \). So the adversarial samples in untargeted attack can be generated by assigning:
\begin{equation}
\label{3.2.3}
\left\{\begin{array}{l}
\mathbf{X}_{0}^{\mathrm{adv}}=\mathbf{X_{\text {ori}}} \\
\mathbf{X}_{n+1}^{\mathrm{adv}}=\mathbf{C}_{\mathrm{X}}^{\epsilon}\left(\mathbf{X}_{n}^{\mathrm{adv}}+\alpha \cdot \operatorname{sign}\left(\nabla_{\mathbf{x}} \mathcal{L}_{PL}(M(\mathbf{X}_{n}^{\mathrm{adv}}), y)\right)
\right)
\end{array}\right.
\end{equation}
 where $C(\cdot)$ is a constraint function that ensures the adversarial samples are in $\epsilon-range$ ball of the clean sample. $y$ is real label $y_{i}^{s}$ for $x_{i}^{s}$ from source domain, and  pseudo label $\hat{y}_{i}^{t}$ for $x_{i}^{t}$ from target domain.
 The process is shown in Fig.\ref{fig:10}.
\subsubsection{Targeted Attack}

Targeted attack tends to add disturbance to the original image, in order to make the  $x_{adv}$ to be predicted to be a selected target label $y_{t}$:
\begin{equation}\label{3.2.4}G(x+r) = y_{t} \end{equation}

Unlike the non-targeted attack, targeted attack will find adversarial perturbations with determined target labels during the learning procedure.
In the targeted attack, we choose the least-likely class according to the prediction of the trained network on image X as the desired target class:
\begin{equation}\label{3.2.5}
y_{t}=\underset{y}{\arg \min }\{p(y \mid \boldsymbol{X})\}
\end{equation}
To acquire an adversarial image which is classified as $y_{t}$, the optimization goal of targeted attack is defined as:
\begin{equation}\label{3.2.6}
\underset{x_{\text {adv }}}{\min } \ \mathcal{L}_{\mathrm{PL}}\left(\mathrm{M}\left(\mathrm{x}_{\text {adv }} ; \theta, \mathrm{w}, \mathrm{b}\right), y_{t}\right)
\end{equation}
We achieve this objective by making iterative steps in the direction of $\operatorname{sign}\left(\nabla_{\mathrm{x}} \mathcal{L}_{P L}\left(M\left(\mathrm{X}_{n}^{\mathrm{adv}}\right), y_{t}\right)\right)$. So the adversarial samples in targeted attack can be generated by:
\begin{equation}
\label{4.2.100}
\left\{\begin{array}{l}
\mathbf{X}_{0}^{\mathrm{adv}}=\mathbf{X}_{\text {ori }} \\
\mathbf{x}_{n+1}^{\mathrm{adv}}=\mathbf{C}_{\mathrm{X}}^{\epsilon}\left(\mathbf{X}_{n}^{\mathrm{adv}}-\alpha \cdot \operatorname{sign}\left(\nabla_{\mathbf{x}} \mathcal{L}_{P L}\left(M\left(\mathbf{X}_{n}^{\mathrm{adv}}\right), y_{t}\right)\right)\right)
\end{array}\right.
\end{equation}

\subsubsection{White-box and Black-box Attack}

\paragraph{White-box attack}

White-box attack requires the attackers to have prior knowledge of the target networks \cite{2016Adversarial,0Boosting}, including the architecture of networks, loss function $\mathcal{L}$ and parameters $\theta,w,b$. It means that the adversarial samples are generated from a network $M^{'}(\theta,w,b)$ who has the exactly the same property, compared with the target network $M(\theta,w,b)$. Sometimes those samples are created by the targeted network directly. In the classification task, the network structure includes both backbones $F(\theta)$ and classifiers $G(w,b)$.

\paragraph{Black-box attack}
Black-box attack means that the attacker cannot obtain the information of the target network. We name the common methods in this scenario \emph{'Random Initialization Method' (RIM)} and \emph{'Avatar Network Method' (ANM)}. For the RIM, \cite{2019Simple} proposes to iteratively select a random direction from a set of specified orthogonal representations, use the confidence degree to check whether it points to or away from the decision boundary, and directly add or subtract vectors to the image to disturb the image. In this way, each update will move the image away from the original image and towards the decision boundary. For the ANM, which is also called transfer attack, we usually train a model $M_{avatar}(\theta^{'},w^{'},b^{'})$, which can be treated as a substituting model of the target model $M(\theta,w,b)$, to generate adversarial samples $X_{adv}$. Then use $X_{adv}$ to attack the target model. The success rate of black-box attack depends heavily on the transferability of adversarial samples. Therefore, the key point of black-box attack based on avatar model is to train an alternative model $M_{avatar}$, whose attributes are as similar to the target model $M$ as possible. Usually, we train the $M_{avatar}$ by using the same datasets as the target model (ground-truth based), or collecting the outputs of the target model(pseudo-label based);

\subsection{Adversarial Domain Defense}

Currently, there is no defense method that works well against all attack methods. This motivates us to first study the properties of state-of-the-art domain adaption networks, and how they affect robustness to various adversarial attacks. A successful defense strategy should meet:

(1) For the clean samples $X_{ori}$, $M_{defense}$ which is trained from randomly initialized model, should have the similar accuracy to the target model $M$;

(2) For the adversarial samples $X_{adv}$, $M_{defense}$ should performs much better than the target model $M$, and even get an accuracy similar to the baseline;

For black-box attack and white-box attack, we benchmark the defense strategies of domain adaption classification task:

\textbf{Defense for white-box attack}. Since the white-box model is completely visible and controllable, our defense strategy is designed as: 1) Learn a target model $M_{target}$ with clean samples $x_{ori}$ and original loss function $L$. This target model $M$ is our attack target. Use $M_{target}$ to generate adversarial sample $x_{adv}$; 2) Mix the clean samples $x_{ori}$ and adversarial $x_{adv}$,  get $x_{mix}$. Then, train a whole new model $M$ with $x_{mix}$ until it converges; 3) Put $x_{ori}$ into the model $M$, update the $x_{adv}$; 4) Repeat 2) and 3) several times until convergence.

Our experimental results show that after several iterations, the defense network $M_{defense}$ has a better performance of classification task on $x_{adv}$. In this sense, we think the model has better robustness to the updated adversarial attack.

\textbf{Defense for black-box attack}. It is meaningless to update adversarial samples of black-box attacks based on transfer attack, since the $x_{adv}$ is generated from $M_{avatar}$, and parameters of $M_{avatar}$ is fixed during defending.  So we mix $x_{adv}$ and $x_{ori}$ to train the target network $M_{defense}$. Such kind of defense method can be regarded as a new data augment method.

\section{EXPERIMENTS AND RESULTS}

\subsection{Datasets and Implementation Details}

    \subsubsection{Datasets.} Office-31 is a mainstream benchmark dataset in visual transfer learning, including 4652 images of 31 categories, which come from three real domains: In this experiment, it is used as a small dataset.

    Office-Home is a benchmark dataset for domain adaptation which contains 4 domains where each domain consists of 65 categories.In this experiment, it is used as a medium scale dataset.

    VisDA-2017 is a simulation-to-real dataset for domain adaptation with over 280,000 images across 12 categories in the training, validation and testing domains.  In this experiment, it is used as a big scale dataset.

   \subsubsection{Adversarial Domain Attack and Defense. }Our main attack method is PL-FGSM, we tried the attack step $\epsilon$ = [0.01, 0.05, 0.1, 0.5] and iteration ITER = 40. Targeted and non-targeted attack can be achieved by  Eq. (\ref{3.2.3} )and Eq. (\ref{4.2.100}). White-box can be achieved by PL-FGSM, and black-box attack can be achieved by PL-FGSM in ANM way described in sec 3.2.1. Besides, Our defense strategy has been described in sec 3.3.2.

   In section 5.2.1, we take non-targeted attack as an example to discuss the influence of different setting of adversarial attack and defense on different models and datasets. In sec 5.2.2, we give the result of non-targeted adversarial to form a comparison with sec 5.2.1.

\subsection{Non-targeted Attack and Defense}

\subsubsection{White-Box Attack. }
\begin{figure*}[t]
  \centering
  \includegraphics[width=\linewidth]{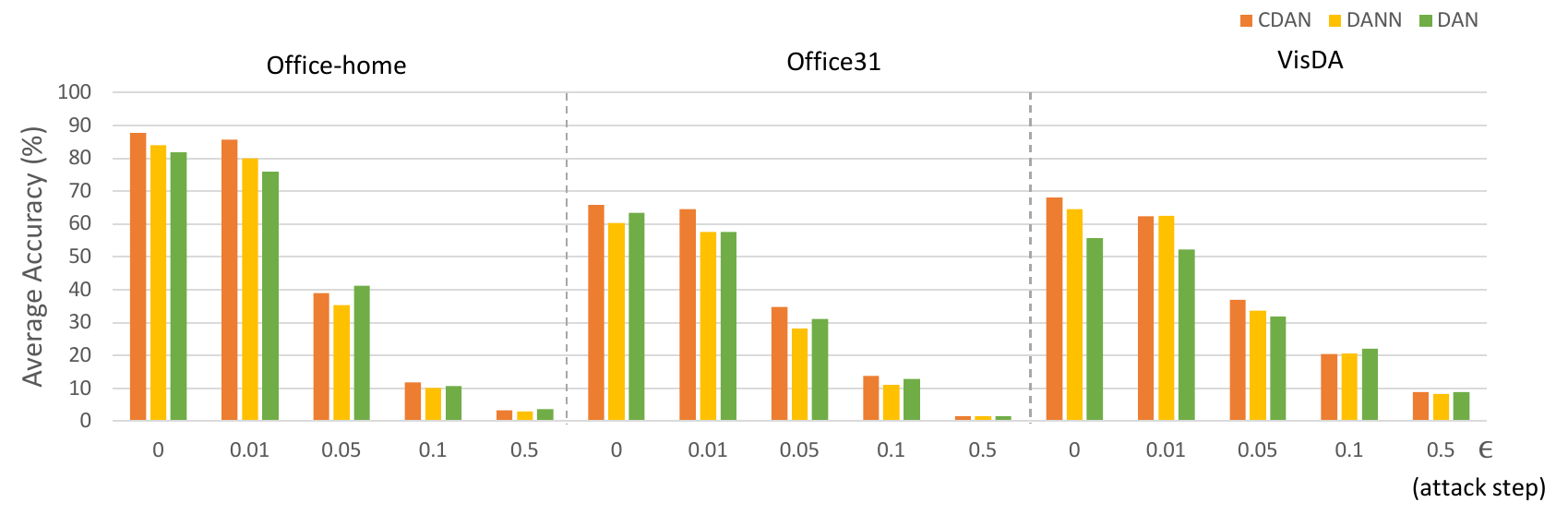}
    \caption{Non-Targeted White-Box Attack. It shows 3 models accuracy after attacked by 4 steps($\epsilon=[0.01, 0.05, 0.1, 0.5]$, and $\epsilon=0$ means the original samples.) on 3 datasets(Office-31, Office-Home, VisDA).}
  \label{fig:2}
\end{figure*}

The attack results of 3 models with four 4 step sizes on 3 datasets are shown in Fig.\ref{fig:2}. The noise does not disturb the image obviously when $\epsilon$ = 0.01. The samples change little at the pixel level, and the performance degradation of the models is very small, which is almost within 10\%. When $\epsilon$ = 0.1 and $\epsilon$ = 0.5, the performance of the model decreases greatly. However, the noises of the images is very large at this time, and the images are already fuzzy. So it is difficult to be used in the actual scene. To this end, We choose the setting of $\epsilon$ = 0.05 to focus on the analysis, where the attack ability is considerable, and the disturbance on the pixel level is acceptable.

The attack accuracy decrease rate results of 3 models on 3 datasets are shown in Fig.\ref{fig:2}.
We compared CDAN to DANN because CDAN is an enhanced DANN model with higher migration performance. On the same dataset, the decrease rate of CDAN is lower than that of DANN, according to the experimental results. It may be deduced that the model's robustness increases as the migration performance improves.
Then we compared DANN and DAN, two basic transfer learning framework models. Because DANN's decline rate is larger than DAN's, we may conclude that an adversarial transfer learning model like DANN is less robust than a methodology like DAN that directly lowers the distribution difference between the two domains.

\subsubsection{White-Box Defense. }

\begin{figure}[h]
  \centering
  \includegraphics[width=\linewidth]{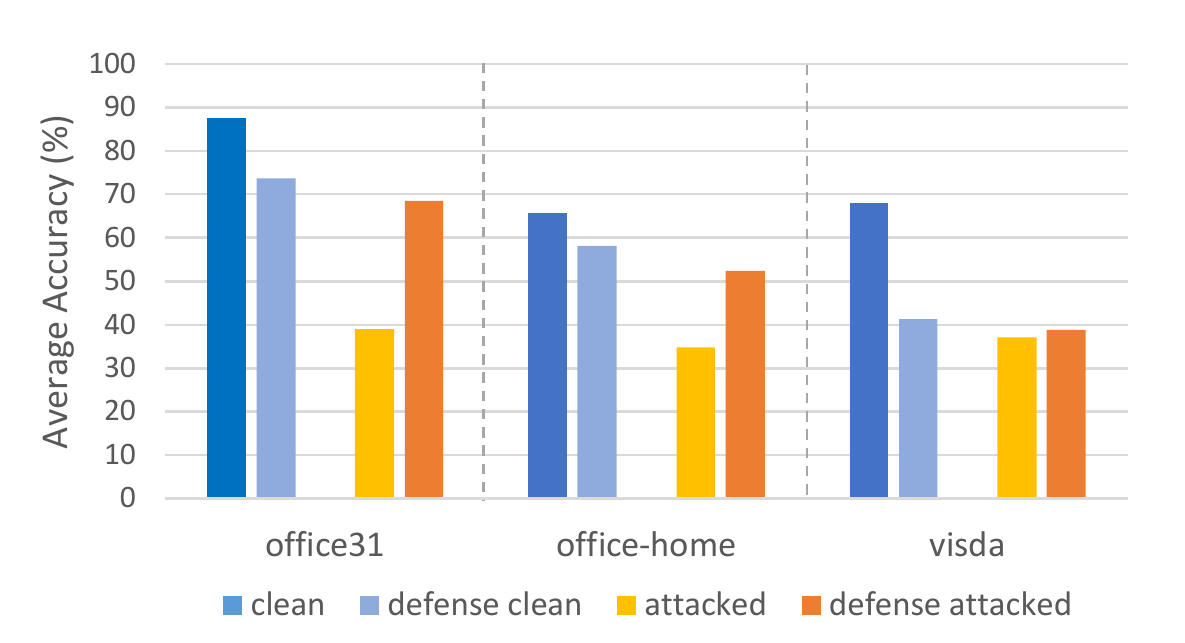}
  \caption{Non-Targeted White-Box Defense. It shows the accuracy of CDAN after white-box defense.}
  \label{fig:3}
\end{figure}

We select CDAN to do white-box defense on 3 datasets. The average results before and after defense are shown in Fig.\ref{fig:3}. After white-box defense,the accuracy of the adversarial samples on The first 2 datasets is greatly improved, but the accuracy of clean samples has been reduced to the same level as that of the adversarial samples.We speculate that the original distribution of the data set will be changed with the addition of confrontation samples, and it will be more difficult to fit the model to the original level. However, the same defense method does not perform well on the VisDA dataset. We speculate that this phenomenon is caused by the sample imbalance between the training set and the test set, which may need some special regularization method.

% Please add the following required packages to your document preamble:
% \usepackage{booktabs}
\begin{table*}[]
\caption{Non-Targeted Black-Box Attack(Office-31). 'Clean' means the accuracy of the clean samples. 'white-box' means the accuracy of the white-box attack shown in fig\ref{fig:1}. 'CDAN2DAN' means use the samples generated by CDAN to attack DAN. 'avarage' means the average accuracy, and this applies to following tables, too.}
\label{table1}\setlength{\tabcolsep}{3.7mm}{
\begin{tabular}{c c c c c c c c c }
%{@{}lllllllll@{}}
\toprule
     & \begin{tabular}[c]{@{}l@{}}Attack\\ source\end{tabular} & AD(\%)    & AW(\%)    & DA(\%)    &DW(\%)    &WA(\%)    & WD(\%) &average(\%) \\ \midrule
DAN  & clean                                                   & 83.73 & 79.62 & 65.70 & 96.85 & 65.56 & 99.79  & 81.88   \\
     & white-box                                               & 40.36 & 48.80 & 34.07 & 57.61 & 19.38 & 46.78  & 41.17   \\
     & DANN2DAN                                                & 41.76 & 39.49 & 44.51 & 66.16 & 40.14 & 53.21  & 47.55   \\
     & CDAN2DAN                                                & 42.36 & 40.88 & 44.44 & 64.40 & 39.75 & 51.80  & 47.27    \\
     \midrule
DANN & clean                                                   & 85.14 & 87.42 & 66.70 & 98.23 & 66.95 & 100.00 & 84.07   \\
     & white-box                                               & 22.69 & 32.32 & 40.64 & 42.76 & 40.43 & 30.12  & 34.83   \\
     & DAN2DANN                                                & 36.04 & 45.65 & 41.67 & 52.76 & 23.35 & 33.33  & 38.80   \\
     & CDAN2DANN                                               & 23.29 & 31.19 & 40.25 & 43.39 & 40.57 & 32.325 & 35.17   \\
     \midrule
CDAN & clean                                                   & 93.17 & 92.32 & 71.31 & 98.61 & 70.64 & 100.00 & 87.67   \\
     & white-box                                               & 28.91 & 37.48 & 44.62 & 45.28 & 32.72 & 45.38  & 39.06   \\
     & DAN2DANN                                                & 38.95 & 53.68 & 46.10 & 60.65 & 19.46 & 47.29  & 44.36   \\
     & DANN2CDAN                                               & 28.71 & 38.23 & 45.01 & 45.28 & 33.26 & 46.18  & 39.44   \\ \bottomrule
\end{tabular}}
\end{table*}

% Please add the following required packages to your document preamble:
% \usepackage{booktabs}
\begin{table*}[t]
\centering
\caption{Non-targeted Black-Box Defense(Office-31).
(attack) means black-box attack average.}
\label{table2}
\setlength{\tabcolsep}{2.5mm}{
\begin{tabular}{c c c c c c c c c c}
\toprule
 & \begin{tabular}[c]{@{}l@{}}Attack\\ source\end{tabular} & samples(\%) & AD(\%) & AW(\%) & DA(\%) & DW(\%) & WA(\%) & WD(\%) & (attack) average(\%) \\ \midrule
DAN  & DANN & clean & 76.14 & 71.56 & 42.89 & 87.63 & 41.26 & 88.11  & (81.88) 67.93 \\
     & DANN & ADV   & 68.48 & 63.48 & 44.01 & 85.66 & 46.39 & 89.26  & (47.55) 66.21 \\
     & CDAN & clean & 82.48 & 65.48 & 47.13 & 91.59 & 43.17 & 93.75  & (81.88) 70.60 \\
     & CDAN & ADV   & 61.45 & 64.41 & 50.42 & 90.68 & 45.47 & 89.48  & (47.27) 66.98 \\
     \midrule
DANN & DAN  & clean & 79.91 & 78.84 & 48.56 & 95.72 & 51.04 & 99.19  & (84.07) 75.53 \\
     & DAN  & ADV   & 59.63 & 64.15 & 50.12 & 87.67 & 37.09 & 88.35  & (38.80) 64.50 \\
     & CDAN & clean & 72.08 & 68.55 & 41.74 & 92.45 & 42.17 & 95.78  & (84.07) 68.79 \\
     & CDAN & ADV   & 65.46 & 66.66 & 49.69 & 92.57 & 46.92 & 96.78  & (35.17) 69.68 \\
     \midrule
CDAN & DAN  & clean & 89.35 & 63.29 & 87.16 & 98.36 & 55.91 & 100.00 & (87.67) 82.34 \\
     & DAN  & ADV   & 77.30 & 61.09 & 81.63 & 93.58 & 49.16 & 97.18  & (44.36) 76.66 \\
     & DANN & clean & 85.14 & 79.25 & 48.84 & 96.47 & 52.28 & 98.99  & (87.67) 76.83 \\
     & DANN & ADV   & 79.71 & 75.97 & 54.70 & 97.61 & 53.28 & 99.79  & (39.44) 76.84 \\ \bottomrule
\end{tabular}}
\end{table*}

\subsubsection{Black-Box Attack. }We discuss the results of the three models' transfer attacks on the Office-31, and $\epsilon$=0.05. The average accuracy of each dataset are showed in Table \ref{table1}. Transfer based attacks result in interesting phenomena: 1)Due to the similar model structure of DANN and CDAN, their mutual attack performance is close. The result of DANN attacking CDAN is 39.44\%, while that of CDAN attacking DANN is 35.17\%. In contrast, due to the large difference between the models, the attack effect of DAN on the other two is poor. This suggests that we may get better results when we attack simple models with complex models, while the attack power of simple models to complex models may be limited; 2)Due to the similarity of CDAN and DANN models, the performance of mutual black-box attack of CDAN and DANN is very close to the result of respective white-box attack. The result of CDAN white-box attack is 39.06\%, and that of DANN attacking CDAN is 39.44\%. It proves the feasibility of black-box attack based on transfer attack. When the avatar network is close to the original network, this black-box attack method can even achieve the performance comparable to the white-box attack.

\subsubsection{Black-Box Defense. } We select $\epsilon$ = 0.05. The defense method refers to the black-box defense described in section 3.2. The results are shown in Table \ref{table2}.

After defense, the accuracy of the model for the adversarial sample is greatly improved, but it is difficult to achieve the accuracy of the original model for clean samples. In general, the average accuracy of the original model for clean samples is more than 80\%, while the accuracy of the trained model for dirty samples is around 60\% - 80\%. Besides, in most cases, the accuracy of defense model to dirty samples is slightly lower than that of clean samples after defense. For example, in the transfer attack of DAN2CDAN,the accuracy of the trained model $F_{defense}$ is 82.34\% for clean samples and 76.66\% for attack samples. This phenomenon will be discussed in sec \ref{subsection5}. Another  interesting thing is that, after the defense training, the accuracy of the defense model for clean samples is slightly reduced, and the decline rate is about 10\%.

\subsection{Targeted Attack and Defense}

By setting this subsection, we compared the targeted attack and defense with non-targeted attack and defense. We select part of models and datasets to repeat the experiment of sec 6.1 in targeted manner.

% Please add the following required packages to your document preamble:
% \usepackage{booktabs}
\begin{table}[H]
\centering
\caption{Targeted White-Box Attack(Office-31)}
\label{table3}
\setlength{\tabcolsep}{2.5mm}{
\begin{tabular}{c c c c c c}
\toprule
    & CDAN(\%)  &  & DANN(\%)  &  & DAN(\%)   \\ \midrule
AD  & 22.28 &  & 21.08 &  & 34.73 \\
AW  & 13.83 &  & 14.71 &  & 30.18 \\
DA  & 32.62 &  & 33.83 &  & 31.16 \\
DW  & 47.16 &  & 45.91 &  & 54.59 \\
WA  & 27.90 &  & 30.35 &  & 18.17 \\
WD  & 29.31 &  & 22.08 &  & 42.36 \\
average & 28.85 &  & 27.99 &  & 35.20 \\ \bottomrule
\end{tabular}}
\end{table}

% Please add the following required packages to your document preamble:
% \usepackage{booktabs}
\begin{table}[H]
\centering
\caption{Targeted White-Box Defense(Office-31)}
\label{table4}
\setlength{\tabcolsep}{0.1mm}{
\begin{tabular}{c c c c c}
\toprule
 &
  clean(\%) &
  \begin{tabular}[c]{@{}l@{}}white-box\end{tabular}(\%) &
  \begin{tabular}[c]{@{}l@{}}$clean_{trained}$\end{tabular}(\%) &
  \begin{tabular}[c]{@{}l@{}}adversarial\end{tabular}(\%) \\ \midrule
AD  & 93.17 & 22.28 & 88.35 & 64.65 \\
AW  & 92.32 & 13.83 & 85.91 & 75.72 \\
DA  & 71.31 & 32.62 & 53.49 & 38.58 \\
DW  & 98.61 & 47.16 & 97.10 & 94.84 \\
WA  & 70.64 & 27.90 & 48.88 & 43.94 \\
WD  & 100.00   & 29.31 & 99.79 & 86.14 \\
average & 87.67 & 28.85 & 78.92 & 67.31 \\ \bottomrule
\end{tabular}}
\end{table}

% Please add the following required packages to your document preamble:
% \usepackage{booktabs}
\begin{table*}[t]
\caption{Targeted Black-Box Attack(Office-31)}
\setlength{\tabcolsep}{4mm}{
\label{table5}
\begin{tabular}{c c c c c c c c c}
\toprule
     & \begin{tabular}[c]{@{}l@{}}Attack\\ source\end{tabular} & AD(\%)    & AW(\%)    & DA(\%)    & DW(\%)    & WA(\%)    & WD(\%)     & average(\%) \\ \midrule
DAN  & clean                                                   & 83.73 & 79.62 & 65.70 & 96.85 & 65.56 & 99.79  & 81.88   \\
     & white-box                                               & 34.73 & 30.18 & 31.16 & 54.59 & 18.17 & 42.36  & 35.20   \\
     & DANN2DAN                                                & 50.20 & 43.14 & 42.52 & 78.92 & 39.97 & 67.67  & 53.74   \\
     & CDAN2DAN                                                & 51.00 & 38.74 & 40.61 & 76.85 & 39.36 & 68.07  & 52.44   \\
     \midrule
DANN & clean                                                   & 85.14 & 87.42 & 66.70 & 98.23 & 66.95 & 100.00 & 84.07   \\
     & white-box                                               & 21.08 & 14.71 & 33.83 & 45.91 & 30.35 & 22.08  & 27.99   \\
     & DAN2DANN                                                & 36.22 & 37.86 & 40.67 & 54.53 & 28.35 & 38.66  & 39.38   \\
     & CDAN2DANN                                               & 22.48 & 15.72 & 31.27 & 47.04 & 33.93 & 23.29  & 28.96   \\
     \midrule
CDAN & clean                                                   & 93.17 & 92.32 & 71.31 & 98.61 & 70.64 & 100.00 & 87.67   \\
     & white-box                                               & 22.28 & 13.83 & 32.62 & 47.16 & 27.90 & 29.31  & 28.85   \\
     & DAN2DANN                                                & 41.55 & 45.86 & 40.14 & 58.40 & 21.64 & 39.33  & 41.15   \\
     & DANN2CDAN                                               & 25.10 & 31.57 & 37.02 & 54.84 & 26.23 & 32.93  & 34.61   \\ \bottomrule
\end{tabular}}
\end{table*}

% Please add the following required packages to your document preamble:
% \usepackage{booktabs}
\begin{table*}[t]
\caption{Targeted Black-Box Defense(Office-31)}
\label{table6}
\setlength{\tabcolsep}{3.0mm}{
\begin{tabular}{c c c c c c c c c c}
\toprule
 & \begin{tabular}[c]{@{}l@{}}Attack\\ source\end{tabular} & samples & AD(\%) & AW(\%) & DA(\%) & DW(\%) & WA(\%) & WD(\%) & (attack) average(\%) \\ \midrule
DAN  & DANN & clean & 76.70 & 75.09 & 56.86 & 96.60 & 58.75 & 98.99  & (81.88) 77.17 \\
     & DANN & ADV   & 66.06 & 62.38 & 48.81 & 93.60 & 51.75 & 95.78  & (53.74) 69.73 \\
     & CDAN & clean & 70.48 & 68.67 & 52.25 & 94.96 & 51.40 & 97.99  & (81.88) 72.62 \\
     & CDAN & ADV   & 60.04 & 58.23 & 46.36 & 91.44 & 46.39 & 93.97  & (52.44) 66.07 \\
     \bottomrule
DANN & DAN  & clean & 77.91 & 77.73 & 45.15 & 97.10 & 50.33 & 97.99  & (84.07) 74.37 \\
     & DAN  & ADV   & 67.47 & 71.95 & 49.91 & 92.32 & 47.78 & 93.17  & (39.38) 70.43 \\
     & CDAN & clean & 72.28 & 75.22 & 38.33 & 94.46 & 46.25 & 98.19  & (84.07) 70.79 \\
     & CDAN & ADV   & 62.44 & 66.28 & 45.61 & 88.05 & 43.02 & 79.51  & (28.96) 64.15 \\
     \bottomrule
CDAN & DAN  & clean & 89.75 & 88.93 & 53.46 & 98.36 & 50.40 & 99.79  & (87.67) 80.12 \\
     & DAN  & ADV   & 77.51 & 86.28 & 55.69 & 97.23 & 52.96 & 98.39  & (41.15) 78.01 \\
     & DANN & clean & 87.14 & 87.04 & 54.34 & 97.61 & 53.67 & 100.00 & (87.67) 79.97 \\
     & DANN & ADV   & 78.31 & 84.15 & 56.44 & 95.97 & 54.31 & 95.38  & (34.61) 77.42 \\ \bottomrule
\end{tabular}}
\end{table*}
We select Office-31 datasets to three models in white-box manner, $\epsilon$=0.05. The attack results are shown in Table \ref{table3}. In general, the  performance of white-box attack is slightly better than that of non-targeted attack, where CDAN decreased by 4.46\%, DANN decreased by 3.59\%, DAN decreased by 2.38\%.

For white-box defense, we choose CDAN model to experiment on Office-31 dataset. The experimental results are shown in Table \ref{table4}, which show similar properties with the defense of non-targeted white-box defense to us: the accuracy of the dirty samples increases greatly, the accuracy of clean samples decreases slightly, and the accuracy of clean samples is always higher than that of dirty samples. Moreover, the accuracy of the model attacked by targeted attack is higher than that of the model attacked by non-targeted method. The former is 78.32\%, the latter is 69.93\%.

For the black box-attack, we also attack each other with the adversarial samples generated by the 3 models, $\epsilon$ = 0.05. The results are shown in Table \ref{table5}. Similarly, the effect of targeted black-box attack is slightly higher than that of targeted white-box attack. For example, for CDAN2DANN, the performance of targeted black-box attack is 1.07\% less than the non-targeted black-box attack. Generally speaking, the result of targeted black-box attack is 1\% - 2\% lower than that of non-targeted black-box attack.

For the non-targeted black-box defense, we chose the same settings as targeted black-box defense, and the results are shown in Table \ref{table6}. Similarly, the accuracy of dirty samples decreases slightly, and the accuracy of dirty samples is improved after defense. But both are difficult to achieve the original accuracy. Similar to non-targeted, the average accuracy of three models after defense is reduced by about 10\%.
\section{DISCUSSION}\label{subsection5}

During the defense of white-box, we find that the accuracy of individual domains is lower than that before defense, i.e. the accuracy of DA's adversarial sample changes from 44.44\% to 43.84\%. Sometimes the test accuracy of clean samples is even lower than that of adversarial samples. We try to control the loss of defense to reduce the overfitting of the model:
\begin{equation}
\label{7.1}
\mathcal{L}=\alpha \mathcal{L}_{\text {cross entropy }}^{a d v}+ \mathcal{L}_{\text {cross entropy }}^{\text {clean }}+\mathcal{L}_{\text {transfer }}
\end{equation}
 where $\mathcal{L}_{\text {cross entropy }}^{a d v}$ and $\mathcal{L}_{\text {cross entropy }}^{clean}$ are the cross entropy losses of adversarial samples and clean samples, and $L_{\text {transfer }}$ is the loss of domain adaption, such as $l_{MMD}$. $\alpha$ is the wight of adversarial samples' loss to trade off the  constraining force of
$\mathcal{L}_{\text {cross entropy }}^{a d v}$ and $\mathcal{L}_{\text {cross entropy }}^{clean}$.
We test the new loss in white-box defense with $\alpha=0.5$(Table \ref{table7}), the accuracy of clean and adversarial samples are both improved, but the test accuracy of clean samples is lower than that of adversarial samples. This problem is worth further discussion.

\begin{table}[H]
\centering
\caption{White-box Defense of weight loss(Office-31)}
\label{table7}
\setlength{\tabcolsep}{1.1mm}{
\begin{tabular}{c c c c c}
\toprule
 &
  clean(\%) &
  \begin{tabular}[c]{@{}l@{}}adv\end{tabular}(\%) &
  \begin{tabular}[c]{@{}l@{}}weight
  clean\end{tabular}(\%) &
  \begin{tabular}[c]{@{}l@{}}weight
  adv\end{tabular}(\%) \\ \midrule
DA  & 45.23 & 43.84  & 50.48 & 51.44 \\
\bottomrule
\end{tabular}}
\end{table}

\section{CONCLUSION}
In this work, we discuss the robustness of the domain adaptation model against adversarial effect. According to the characteristics of existing domain adaptation methods, we propose a fast gradient descent method based on pseudo label(PL-FGSM), which can be used as a basic method of adversarial attack for domain adaption. Experiments show that the performance of the existing models is greatly degraded under PL-FGSM's attacking. This shows the limitation of cross domain model's robustness.

In order to stimulate the community's research on the above problems, we benchmark the various adversarial settings in domain adaptation, including white-box and black-box attack, targeted and non-targeted attack,ect. Through extensive experiments of three models on three datasets in small, medium and large scale, we get some interesting phenomena and meaningful conclusions. We hope that these results can provide a meaningful reference for future work.In addition, we discussed some of the unique properties of cross domain defense. Our code is open source and available to promote the development of cross domain attack and effective defense.

\bibliographystyle{ACM-Reference-Format}
\bibliography{sample-base}
%%
%% If your work has an appendix, this is the place to put it.
\appendix

\end{document}